\pdfoutput=1

\documentclass[11pt]{article}

\usepackage[preprint]{acl}

\usepackage{times}
\usepackage{latexsym}

\usepackage[T1]{fontenc}

\usepackage[utf8]{inputenc}

\usepackage{microtype}

\usepackage{inconsolata}

\usepackage{graphicx}

\newcommand{\review}[1]{{\textcolor{black}{#1}}}

\usepackage{hyperref} 
\usepackage{bm}
\usepackage{enumitem} 
\usepackage{multirow} 
\usepackage{booktabs} 
\usepackage{longtable}
\usepackage{soul}

\title{LLM-driven Constrained Copy Generation through Iterative Refinement}

\author{\textbf{Varun Vasudevan}, \textbf{Faezeh Akhavizadegan}, \textbf{Abhinav Prakash}, \textbf{Yokila Arora}, \\ \textbf{Jason Cho}, \textbf{Tanya Mendiratta}, \textbf{Sushant Kumar},  \textbf{Kannan Achan} \\ Walmart Global Tech \\ Sunnyvale, CA, USA -- 94086\\
\{varun.vasudevan, faezeh.akhavizadegan, abhinav.prakash, yokila.arora, \\ jason.cho, tanya.mendiratta, sushant.kumar, kannan.achan\}@walmart.com}

\begin{document}
\maketitle
\begin{abstract}
Crafting a marketing message (copy), or copywriting is a challenging generation task, as the copy must adhere to various constraints. Copy creation is inherently iterative for humans, starting with an initial draft followed by successive refinements. However, manual copy creation is time-consuming and expensive, resulting in only a few copies for each use case. This limitation restricts our ability to personalize content to customers. Contrary to the manual approach, LLMs can generate copies quickly, but the generated content does not consistently meet all the constraints on the first attempt (similar to humans). While recent studies have shown promise in improving constrained generation through iterative refinement, they have primarily addressed tasks with only a few simple constraints. Consequently, the effectiveness of iterative refinement for tasks such as copy generation, which involves many intricate constraints, remains unclear. To address this gap, we propose an LLM-based end-to-end framework for scalable copy generation using iterative refinement. To the best of our knowledge, this is the first study to address multiple challenging constraints simultaneously in copy generation. Examples of these constraints include length, topics, keywords, preferred lexical ordering, and tone of voice. We demonstrate the performance of our framework by creating copies for e-commerce banners for three different use cases of varying complexity. Our results show that iterative refinement increases the copy success rate by 16.25--35.91\% across use cases. Furthermore, the copies generated using our approach outperformed manually created content in multiple pilot studies using a multi-armed bandit framework. The winning copy improved the click-through rate by 38.5--45.21\%.
\end{abstract}

\section{Introduction}

\begin{figure*}
\centering
\includegraphics[scale=0.28]{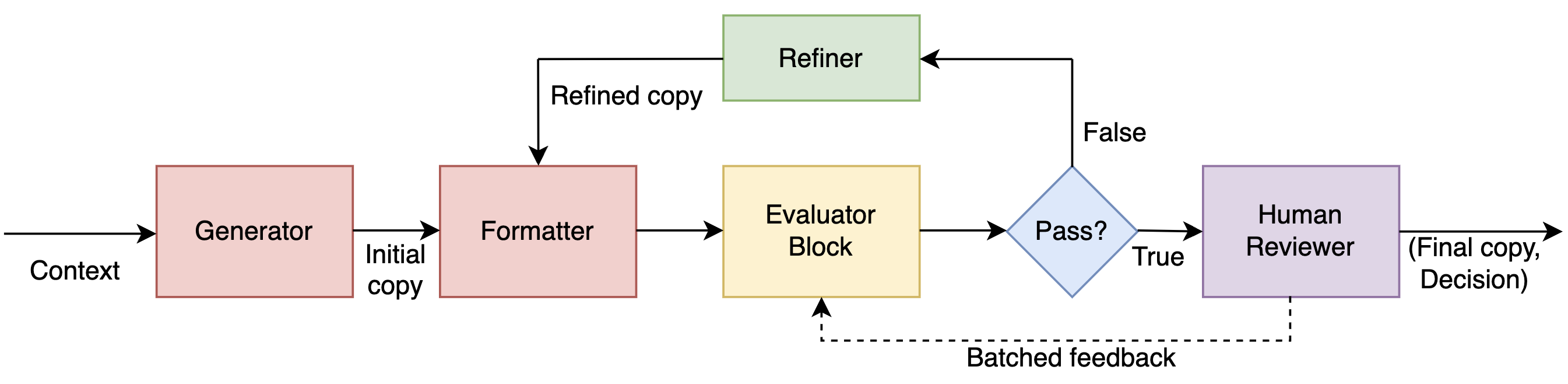}
\caption{LLM-based end-to-end framework for scalable copy generation. It has three main blocks: a generator, an evaluator, and a refiner. The generated copy undergoes formatting and evaluation. If any evaluation fails, it is sent to the refiner for revision. This process continues until the copy passes all evaluations or the maximum number of refine attempts is reached. Once a copy passes all evaluations, it goes to human reviewers.}
\label{fig:architecture}
\end{figure*}

E-commerce platforms are home to hundreds of digital banners that change periodically. A banner typically includes a visually appealing image and text conveying a marketing message, aka copy. Traditionally, copywriters manually create the copy, adhering to predefined constraints. This manual curation process is expensive and time-intensive. Therefore, only a few copies ($N{<}10$) are created for each banner. This limitation restricts our ability to personalize banners for different customer cohorts. Personalized copies can help businesses meet individual customer needs, improve customer engagement, and ultimately drive revenue \citep{chen2024large, yang2023palr, chandra2022personalization}.

In the field of text generation, the terms ``controlled text generation'' \citep{zhou2023controlled, sun2023evaluating} and ``controllable text generation'' \citep{zhang2023survey} are often used interchangeably to refer to tasks that generate text with predefined constraints. Given the overlap and subtle nuances in interpreting these terms, we will adopt the term ``constrained text generation''  to refer to such tasks.

Practitioners are increasingly adopting large language models (LLMs) for creative constrained text generation tasks such as copywriting \citep{zhou2024gcof, wang2024weaver} and storytelling \citep{xie2023next, simon2022tattletale}.  This paper focuses on using LLMs for copywriting for e-commerce banners. Generating the image component of e-commerce banners using text-to-image stable diffusion models is another area of active research, which is beyond the scope of this paper \citep{vashishtha2024chaining, yang2024new}. 

Unlike humans, LLMs can quickly generate hundreds of copies, but the generated content does not consistently meet all the constraints (verbalized as instructions in the prompt) on the first attempt. \citet{zhou2023instruction} find that the instruction-following accuracy of an LLM varies depending on the type of instruction. In the case of copywriting for e-commerce banners, the generated copy must satisfy many intricate constraints simultaneously. These include, but are not limited to, adherence to a specified length, the inclusion of specific keywords and topics aligned with the banner's intent, preferred lexical ordering, and conformity to the brand's voice and tone while maintaining a high degree of readability and engagement. Such constraints make copywriting for e-commerce banners a significantly harder task than the lightly constrained tasks explored in \citet{madaan2024self}, \citet{sun2023evaluating}, and \citet{zhou2023instruction}.

Inspired by the recent work of \citet{madaan2024self} on iterative self-refinement, we propose an LLM-based end-to-end framework for scalable copy generation through iterative refinement. The proposed framework shown in Figure~\ref{fig:architecture} has three major components: a generator, an evaluator block (a set of evaluators), and a refiner. The generator creates an initial copy for a given use case. Each evaluator in the evaluator block assesses the generated copy on a single constraint providing an evaluation decision and a feedback. 
Any copy that fails an evaluation is immediately passed to the refiner along with its feedback for revision. This process continues until the copy passes all the evaluations or a maximum number of refine attempts are reached. Our key contributions are as follows.
\setlist{nolistsep}
\begin{itemize}[noitemsep]
\item We propose a practical and efficient framework for LLM-driven copy generation through iterative refinement. This framework can simultaneously handle many intricate constraints of copy generation and can successfully generate a large number of diverse personalized copies. Prior studies on copywriting and iterative refinement primarily focus on fewer and simpler constraints.
\item We demonstrate the performance of our framework on three challenging use cases of varying complexity. Our results show a significant improvement in success rate across all use cases compared to the base LLM.
\item \review{We showcase the business impact of our framework in improving user engagement via multiple pilot studies, which resulted in a significant lift in CTR compared to manually created copies.}
\end{itemize}

\begin{table}
\centering
\small
\begin{tabular}{l}
\toprule
Paper: Constraints\\
\midrule 
\citet{guo2022automatic}, \citet{lin2024generating}, \citet{wei-etal-2022-creater}: topic\\ 
\citet{zhou2024gcof}: keyword, length \\
\midrule
\textbf{Ours}: keyword, topic, length, tone, lexical ordering, etc. \\
\bottomrule
\end{tabular}
\caption{Constraints in related copywriting literature.}
\label{tab:related-work-constraints}
\end{table}

\section{Related work}

\textbf{Constrained text generation.} \citet{ashok2024controllable} show that prompting instruction-tuned LLMs to achieve constrained generation outperforms the use of methods such as FUDGE \citep{yang-klein-2021-fudge}, NeuroLogic decoding \citep{lu2022neurologic}, and DEXPERTS \citep{liu2021dexperts} that attempt to steer the generation process at the token level. Our paper focuses on constrained generation that verbalizes constraints as instructions in the prompt, and improves the generation iteratively.

\citet{zhou2023instruction} show that the instruction-following accuracy of an LLM depends on the type of instruction. For instance, GPT-4 has instruction-following accuracies of approximately 85\%, 73\%, and 68\% for instructions on keywords, length, and punctuation, respectively. In another work, \citet{sun2023evaluating} show that LLMs struggle with fine-grained numerical constraints (e.g., word count in generated text) but mostly conform to coarse constraints such as keyword, sentiment, and topic incorporation. The copy generation task discussed in this paper has all the aforementioned constraints and more, making it significantly challenging for LLMs to meet on the first attempt.

\textbf{Copywriting.} \citet{guo2022automatic} and \citet{lin2024generating} fine-tune Transformer based models for product copywriting. It involves writing accurate and engaging descriptions for products. Typically existing corpus of copy and customer reviews are used as data for training. In a similar line of work, \citet{wei-etal-2022-creater} propose a fine-tuned model for generating copy for items based on user reviews. Their model learns from online A/B test data to maximize click-through rate (CTR). Another related work is GCOF \citep{zhou2024gcof}, where authors use LLMs to write copy for ads using a list of keywords. \autoref{tab:related-work-constraints} compares the constraints in prior work with our study. Fine-tuning an LLM is impractical for us due to data scarcity, the changing nature of constraints from one use case to another, and evolving business needs that drive changes in constraints. Given the significance of creative content generation and its impact on various businesses, companies are also developing foundational models for these tasks. \citep{wang2024weaver}.

\textbf{Iterative self-refinement.} \citet{madaan2024self} demonstrate that LLM output improves through iterative self-refinement by providing specific, actionable feedback to the LLM. However, the constrained generation task explored in their paper is relatively simple, with just keyword constraints. \citet{jiang2024self} hypothesize that for LLMs to continually self-improve their previous outputs, they should be better at discriminating previous outputs than generate initial responses. On testing their hypothesis on the constrained generation task in \citet{madaan2024self}, \citet{jiang2024self} found that LLMs showed only marginal preference (54.7\%) for self-refined outputs. In a similar line of work, \citet{huang2023large} demonstrate that current LLMs cannot improve their reasoning performance through intrinsic self-correction, and recommend using external feedback whenever available to improve model performance. In light of these prior works, the effectiveness of iterative refinement for tasks such as copy generation, which involves many intricate constraints, is unclear.

\textbf{LLMs for evaluation.} Using LLMs to grade copy can significantly reduce the time and cost associated with human evaluation \citep{wei2022chain, lee2024checkeval, zheng2024judging}. Similar to the approach in CheckEval \citep{lee2024checkeval}, each evaluator in the evaluator block grades the copy on a single aspect. The LLM-based evaluators in our framework use chain-of-thought prompting \citep{wei2022chain} and ``Think step-by-step'' strategy for reasoning \citep{kojima2022large}. The design of these evaluators also draws inspiration from the insights and conclusions in \citet{sclar2023quantifying}, \citet{huang2023large}, and \citet{laban2023you}.  

\section{Proposed framework: Copy generation through iterative refinement}

\subsection{Problem definition} 
A copy is crafted for a particular use case for a certain audience, conveying a specific marketing message (topic). The structure and length of a copy depends on the use case. For example, a copy may have a single string or a combination of two or more coherent and connected strings, such as a header and subheader. All copies must adhere to brand guidelines such as tone of voice and punctuation requirements, avoid using off-brand words, and be able to communicate the value proposition to the intended audience. The guidelines and use cases also evolve over time, bringing in new requirements.    

\subsection{Generator and formatter}
The generator block consists of an input prompt $x_c$ for a given context $c$ and an LLM $G$ characterized by parameters $\theta$. The context specifies the use case and is used to query relevant constraints. The prompt consists of \textit{role}, \textit{contextual descriptions}, \textit{instructions}, \textit{usecase-specific instructions}, and \textit{examples}. The \textit{role} component designates the LLM as a creative copywriter and sets the high-level objective of the task \citep{shanahan2023role}, whereas the \textit{contextual descriptions} supply context-specific information, such as defining a user cohort. The \textit{instructions} component provides instructions on tone and brand constraints that do not change across use cases, and the \textit{task-specific instructions} include specific requirements of the task followed by a few \textit{examples} for in-context learning when available. 

The generation process begins with the creation of an initial copy: \(y_c^{(0)} = G(x_c; \theta),\) which is then passed along with the context $c$ through a formatter $P$. The formatter starts by parsing the output into a valid \texttt{JSON} string with \texttt{header} and/or \texttt{subheader} keys as required by the use case (see Table~\ref{tab:tasks}).
The parsed copy is then modified using predefined rules to meet necessary criteria, producing an output: \(\tilde{y}_c^{(0)} = P(y_c^{(0)}, c).\) Few examples of formatting include changing ``and'' to ``\&'', removing serial commas, and stripping punctuation from the end of single-sentence copies.

We adopt a batch generation approach to efficiently handle the requirement of generating many copies for a use case. We produce $m$ (typically 10--20) copies per batch rather than generating them individually. This is done by adding an instruction in the prompt to generate $m$ different copies. The $m$ copies are then individually formatted, evaluated, and refined.

\subsection{Evaluator}
The evaluator block uses deterministic Python functions for assessing simple constraints like the length of a copy, and LLM-based evaluators for more nuanced aspects such as tone of voice. LLM-based evaluators incorporate chain-of-thought and few-shot learning techniques to improve performance. Some evaluators, such as the length checker, grade each component of the copy (\texttt{header}/\texttt{subheader}) separately, while others, such as the coherence checker, grade the whole copy.

Let $\mathcal{E}$ be the set of all available evaluators. Given a context $c$, we query a sequence of evaluators, $\bm{E}_c {=} (E_1, E_2, \cdots, E_{n_c})$ where $E_i \in \mathcal{E}$ for all $i \in 1,\cdots,n_c$. The subset and the sequence of evaluators are predetermined by the context. Each evaluator returns a binary value $s_i$ (1: pass, 0: fail) and a structured feedback string $f_i$ with the rejection reason. $f_i$ is set to a default value when $s_i {=} 1$. 

The formatted copy $\tilde{y}_c^{(0)}$ and the context $c$ is passed through the sequence of evaluators $\bm{E}_c$ until one of the following conditions is met:  
\setlist{nolistsep}
\begin{itemize}[noitemsep]
\item $s_i {=} 1$ for all $i$. It indicates the copy passed all the evaluations. Consequently, the copy is forwarded to a team of copywriters and legal professionals for a final human-in-the-loop evaluation before placing on our site. If a copy is rejected at this stage, the feedback is logged and used to improve the evaluator and refiner blocks for subsequent copy generations. 
\item $s_i {=} 0$, indicating evaluator $E_i$ failed the copy. Consequently, the copy $\tilde{y}_c^{(0)}$, context $c$, and feedback $f_i$ are passed to the refiner. 
\end{itemize}

\subsection{Refiner} The refiner block consists of an LLM $G^\prime$ and a prompt generator. We use the same LLM for generation and refinement. The prompt generator maps the evaluation feedback $f_i$ to an actionable (clear and specific) instruction. It also retrieves relevant \textit{contextual descriptions} using the context $c$ to form the complete refiner prompt $z_{c,i}$. The refiner prompt $z_{c,i}$ and the copy $\tilde{y}_c^{(j)}$ are passed to $G^\prime$ to generate a refined copy:
 \(y_c^{(j+1)} = G^\prime(z_{c,i} \Vert \tilde{y}_c^{(j)}),\)
where $\Vert$ denotes string concatenation, $j {<} J$, and $J$ is the maximum number of refine attempts. $J$ is typically set to 1 or 2. If a copy doesn't pass all the evaluations even after $J$ refine attempts, then it is discarded. See Figure~\ref{fig:message_flow} for an example of how the feedback (hyperbolic terms) from a tone of voice evaluator is used to refine the copy.

\begin{table}
\centering
\small
\begin{tabular}{l c c}
\toprule
Use case    & Copy structure  & length constraint $l {\leq} L$ \\
\midrule 
\texttt{Campaign-A}  & \texttt{header}  & $L \in [40,60]$ \\
\midrule
\multirow{2}{*}{\texttt{Campaign-B}} & \texttt{header} & $L \in [30,40]$ \\ 
& \texttt{subheader} & $L \in [60,80]$ \\
\midrule
\multirow{2}{*}{\texttt{Campaign-C}} & \texttt{header} & $L \in [80,100]$ \\ 
& \texttt{subheader} & $L \in [80,100]$ \\
\bottomrule
\end{tabular}
\caption{Three use cases of varying complexity and their structure and length constraints.}
\label{tab:tasks}
\end{table}

\section{Use cases and evaluation criteria}

\subsection{Use cases} 
\label{subsec:use_cases}
\review{We are a retail company with both a physical and online presence. We offer multiple services to our customers, such as tire maintenance, refill prescriptions, free shipping of items without order minimum, free delivery of groceries and other essentials from nearby stores, video streaming service, and fuel discounts at participating gas stations. We have banners promoting these services on different pages on our website.} Each banner is for a different use case. The proposed framework has been deployed for three use cases listed in Table~\ref{tab:tasks}. Their description is as follows.

\review{\texttt{\textbf{Campaign-A}} aims to create awareness about our services to customers who haven't used them before.} Each copy is a single string with a length constraint and promotes a single service (single-topic generation). \review{\texttt{\textbf{Campaign-B}} aims to re-engage customers with services they have used before.} It is a single-topic generation task like \texttt{Campaign-A}. However, each copy has two strings, a \texttt{header} and \texttt{subheader}, each with its length constraint. The \texttt{header} and \texttt{subheader} should together form a coherent message. \texttt{\textbf{Campaign-C}} aims to encourage customers who have not used our services to try them for free for a limited period. Copy structure is the same as \texttt{Campaign-B}. However, each copy promotes more than one service (multi-topic generation) and is personalized to a specific user cohort, such as pet owners and people with family. Examples for few constraints from the three use cases described above is given in \autoref{sec:constraints}.

\subsection{Evaluation Criteria} 
We evaluate the proposed framework on: 
\setlist{nolistsep}
\begin{enumerate}[noitemsep]
\item \textbf{The effectiveness of iterative refinement.} Following \citet{sun2023evaluating}, we define the success rate for copy generation as the percentage of generated copies that pass all the evaluations. The change in success rate with iterative refinement denotes its effectiveness.
\item \textbf{The effectiveness of generated copies on user engagement.} It is done by measuring the CTR of the generated copies in pilot studies. To test multiple copies together and increase web traffic allocation to better-performing copies over time, we deploy a multi-armed bandit (MAB) framework \citep{silva2022multi}.
\end{enumerate}

\section{Results and Discussion}
\subsection{Effectiveness of iterative refinement} Table~\ref{tab:success_rate_overall} shows the success rates with and without iterative refinement for all three use cases. On average, the success rate increased by 23, 34.07, and 16.25 percentage points for campaigns \texttt{A}, \texttt{B} and \texttt{C}, respectively. For campaigns \texttt{A} and \texttt{B}, we have shown results for two services: free delivery from the store and free shipping without order minimum. In both cases, the success rate for free shipping service is lower because the brand guideline constraints for this service are more complex than those for free delivery service. Table~\ref{tab:campaign-A-examp} shows a few examples of copy from \texttt{Campaign-A}. Note the diversity in the choice of words and the style of communication among copies of the same service. Refer \autoref{sec:appendix-campaignB} for examples of copies from \texttt{Campaign-B}.

LLMs struggle with length constraints as noted in prior works \citep{sun2023evaluating, zhou2023instruction}. When considering the length constraint, \texttt{Campaign-B} is the hardest with its short header. It is reflected in the number of generated copies that violated the length constraint: 72.16\% of the failures for the free delivery service and 56.86\% of the failures for the free shipping service. After refinement, on average, 59.31\% of those copies passed all evaluations.

\begin{figure*}[!htb]
\centering
\includegraphics[width=\textwidth]{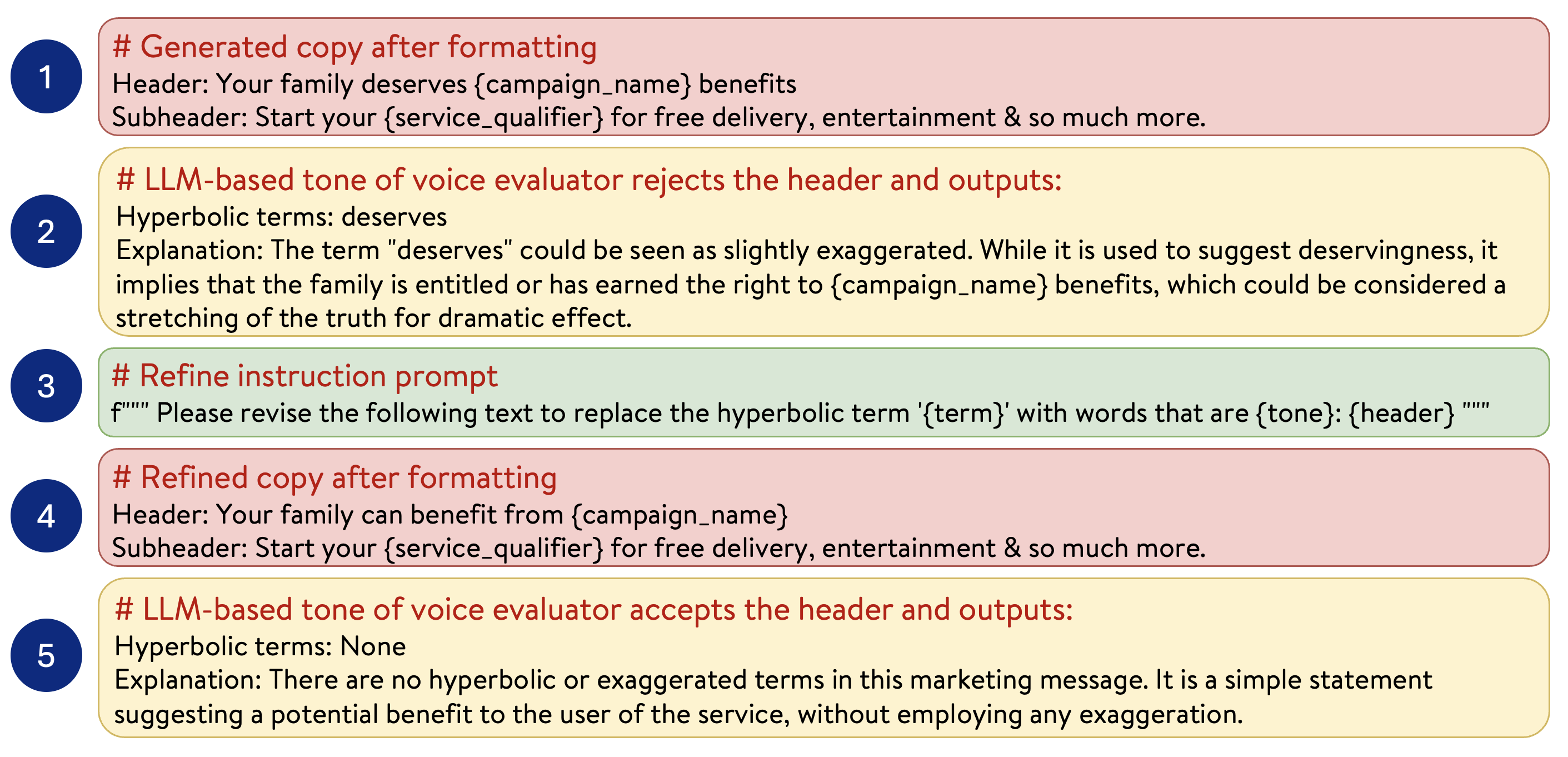}
\caption{\review{Example of iterative refinement. The generated copy in (1) is rejected by a tone of voice evaluator with feedback shown in (2). This feedback is used in the refine prompt in (3) to get the refined copy in (4). The tone of voice evaluator accepts the refined copy with feedback shown in (5). \{tone\} contains tone of voice keywords. \{campaign\_name\} is the name for the campaign and \{service\_qualifier\} is any specific attribute such as ``free trial''.}}
\label{fig:message_flow}
\end{figure*}

In \texttt{Campaign-C}, we generate copies for five different user cohorts or persona \citep{persona_model}. Figure~\ref{fig:message_flow} illustrates the flow of a sample copy (directed at families) through our framework. The \texttt{header} in this copy is rejected by a tone of voice evaluator in the first round of evaluation and subsequently refined. The refined copy passes the tone of voice evaluator.

\textbf{Ablation study.} To study the effect of adding a new constraint to a task, we take the \texttt{Campaign-A} use case and add the constraint that the copy should now target a specific persona. The results is shown in Table~\ref{tab:acq-persona}. The success rate without iterative refinement after adding the new constraint is 24.76\%, roughly 16\% lower than the success rate without the constraint (see Table~\ref{tab:success_rate_overall}). As seen in Table~\ref{tab:acq-persona}, 38.6\% (122/316) of the failed copies were rejected by the persona evaluator. For brevity, we show results after refinement only for those 122 copies. Upon refinement, while all 122 copies passed the persona evaluator, only 46.72\% (57/122) of them passed all evaluations. A human reviewer validated the correctness of LLM-based persona evaluation on the 122 copies before and after refinement. 

\begin{table}
\centering
\small
\begin{tabular}{lcc}
\toprule
& \multicolumn{2}{c}{\textbf{Success rate (\%)}} \\
\textbf{Use case}    & Base LLM & Base LLM + Refine\\
\midrule 
\texttt{\textbf{Campaign-A}} & & \\
Free delivery ($N{=}100$) & 41.0 & 65.0 ($\uparrow 24.0$)\\
Free shipping  ($N{=}100$) & 34.0 & 56.0 ($\uparrow 22.0$)\\
\midrule 
\texttt{\textbf{Campaign-B}} &  & \\
Free delivery ($N{=}180$) & 46.11 & 78.33 ($\uparrow 32.22$)\\
Free shipping ($N{=}220$) & 30.45 &  66.36 ($\uparrow 35.91$)\\
\midrule
\texttt{\textbf{Campaign-C}} ($N{=}437$)  & 41.19 & 57.44 ($\uparrow 16.25$)\\
\bottomrule
\end{tabular}
\caption{Iterative refinement significantly increases the success rate. $N$ is the number of samples.}
\label{tab:success_rate_overall}
\end{table}

\begin{table}
\centering
\small
\begin{tabular}{l}
\toprule
\textbf{Service: Free delivery from store}\\
\midrule 
\review{Free delivery from stores saves you time \& money} \\
\review{Keep your kitchen stocked with free delivery from stores} \\ 
\review{Shop from the comfort of home with free delivery} \\
\review{Free delivery from stores: the ultimate time saver} \\
\toprule 
\textbf{Service: Free shipping with no order minimum}\\
\midrule 
\review{This is not a drill: get free shipping with no order minimum} \\
\review{Don't miss out: get free shipping with no order minimum} \\
\review{Guess what? There's no minimum order for free shipping.} \\
\review{Free shipping with no order minimum. You read it right.} \\
\bottomrule
\end{tabular}
\caption{Qualitative presentation of few copies generated by our approach for \texttt{Campaign-A}. Observe the diversity among copies of the same service.}
\label{tab:campaign-A-examp}
\end{table}

\begin{table}[!htb]
\centering
\small
\begin{tabular}{lc}
\toprule
\textbf{Description}   & \textbf{Count} \\
\midrule 
Generated messages & 420 $(A)$ \\
Accepted by Evaluator (no refinement) & 104 $(B)$ \\
\textbf{Success rate without refinement} $B/A$ & 24.76\% \\
\midrule 
\textbf{Rejected by Evaluator block} & 316 \\
- Rejected by Persona evaluator (out of 316)  & 122  \\
\midrule 
\textbf{Status of 122 copies after refinement} &  \\
- Rejected by Persona evaluator & 0  \\
- Failed length constraint & 21  \\
- Failed value proposition & 23 \\
- Failed tone of voice & 21 \\
- Total failed after refinement & 65 \\
- Total passed after refinement & 57 $(C)$ \\
\midrule 
\textbf{Success rate after refinement} $(B$+$C)/A$ & 38.33\%\\ 
\bottomrule
\end{tabular}
\caption{Breakdown of ablation study results after adding a user cohort constraint to \texttt{Campaign-A}.}
\label{tab:acq-persona}
\end{table}

\subsection{Impact on user engagement} 
We conducted pilot studies on our e-commerce platform for campaigns \texttt{A} and \texttt{B} using MAB. In both studies, the winning variant was a copy generated through our approach. Results showed an increase in CTR by 38.5\% in \texttt{campaign-A} and 45.21\% in \texttt{campaign-B} for the winning variant compared to manual content. It demonstrates the engagement of the generated content.

\subsection{Practical considerations}
The following are a few practical considerations to keep in mind when deploying a copy generation system like ours.
 
\textbf{Diversity.} When we generate a large number ($N{>}50$) of copies, a small fraction of them are duplicates or copies with only minor differences. To address this, we use a diverse subset selection algorithm \citep{mahabadi2023improved} to identify and shortlist $k{<}N$ distinct copies.

\textbf{Evaluation criteria.} It is infeasible to enumerate all possible evaluation criteria to grade LLM-generated copy. Human judges refine existing evaluation criteria or introduce new criteria as they grade more LLM outputs. This phenomenon was first reported in a user study by \citet{shankar2024validates} as ``criteria drift.'' The drift happens because humans are exposed to new types of poor LLM outputs as they grade more. So, as we scale our evaluator block is also evolving continually.


\section{Conclusion}

Prior work on constrained generation using iterative refinement and copy generation using LLMs focus on a few simple constraints. Therefore, the effectiveness of iterative refinement for industrial scale copy generation tasks with many intricate constraints was unclear. To address this gap, we proposed an LLM-based end-to-end framework for scalable copy generation using iterative refinement. Our framework has three major components: a generator, an evaluator block (a set of evaluators), and a refiner. Our results show that iterative refinement increases the success rate by generating copies that satisfy all constraints. We also observed a significant lift in CTR when the generated copies were deployed on our platform. To the best of our knowledge, this is the first study to address multiple challenging constraints simultaneously in copy generation.

\bibliography{main}
\newpage
\appendix

\section{Constraints}
\label{sec:constraints}
\autoref{tab:example-constraints} on the next page shows examples of few constraints from the the copy generation use cases described in this paper. This list of constraints is not exhaustive.

\begin{table*}
\centering
\begin{tabular}{lll}
\toprule
\textbf{S.No.} & \textbf{Constraint Type} & \textbf{Constraint}\\
\midrule 
1 & Length & Should be less than \{N\} characters. \\
2 & Topic inclusion & Should talk about \{service name\}. E.g. free delivery from store.\\
3 & Topic exclusion & Should not contain info that is also present in \{call to action\}.\\ 
4 & Tone of voice & Tone of copy should be: \{tone keywords and their description\}. \\
5 & Style & Style of copy should be: \{style\}. E.g. assertive, question-answer. \\
6 & Keyword inclusion & Should contain the keywords: \{list of keywords\}. \\
7 & Keyword exclusion & Should not contain these off-brand words: \{list of off-brand words\}. \\ 
8 & Punctuation & No punctuation after the following words: \{word list\}. \\
9 & Lexical ordering & Prefer the term \{A\} instead of \{B\}. \\
10 & Coherence & Header and subheader should form a coherent message. \\
\bottomrule
\end{tabular}
\caption{Examples of few constraints from the copy generation task described in this paper. This list is not exhaustive.}
\label{tab:example-constraints}
\end{table*}

\section{Examples of copy for campaign-B}
\label{sec:appendix-campaignB}
Few copies generated using our framework for \texttt{Campaign-B} are shown in \autoref{tab:campaignB-examples} on the next page. Each copy has two components, a header and a subheader. 
\begin{table*}
\centering
\begin{tabular}{l}
\toprule
\textbf{Service: Free delivery from store}\\
\midrule 
\textbf{Header:} Leave the store trip to us \\
\textbf{Subheader:} Free \& fast delivery directly from your local store. \\
\midrule
\textbf{Header:} Shop from home, save more time \\ 
\textbf{Subheader:} \{campaign name\} brings groceries \& more right to your home with free delivery. \\ 
\toprule 
\textbf{Service: Free shipping with no order minimum}\\
\midrule 
\textbf{Header:} Maximize your savings\\
\textbf{Subheader:} Enjoy free shipping with no order minimum. Only with \{campaign name\}. \\
\midrule
\textbf{Header:} \{campaign name\} unlocks free shipping\\
\textbf{Subheader:} No order minimum means more savings, more often. \\
\bottomrule 
\end{tabular}
\caption{Few copies generated using our framework for \texttt{Campaign-B}.}
\label{tab:campaignB-examples}
\end{table*}

\end{document}